\documentclass[runningheads]{llncs}

\usepackage[a4paper,top=2.5cm,bottom=2.5cm,left=3.1cm,right=3.1cm,marginparwidth=2cm]{geometry}

\usepackage[german=quotes]{csquotes}
\usepackage{amsmath}
\usepackage{graphicx}
\usepackage{booktabs}
\usepackage{pdflscape}
\usepackage{afterpage}
\usepackage{longtable}
\usepackage{amssymb}
\usepackage{mathtools}
\usepackage{xurl}
\usepackage{siunitx}
\usepackage{textcomp}
\usepackage[table]{xcolor} % Required for coloring table elements
\usepackage{multirow} % Required for multirow cells
\usepackage{siunitx}
\usepackage{textcomp}

\usepackage[
colorlinks=true,
citecolor=teal,
urlcolor=violet,
linkcolor=purple,
draft=false,
]{hyperref}

\usepackage[
    textsize=footnotesize,
    ]{todonotes}

\usepackage[
noabbrev,
capitalize,
]{cleveref}

\usepackage[
style=alphabetic,
backend=biber,
]{biblatex}
\newcommand{\rsquared}[0]{\(\text{R}^2\)}

\begin{document}
	
	\title{The Applicability of Federated Learning to Official Statistics}
	%\author{Anonymous author(s)} % für anonymes Peer-Review
	\author{Joshua Stock\inst{1} \and Oliver Hauke\inst{2} \and Julius Wei\ss{}mann\inst{2} \and Hannes Federrath\inst{1}}
	\institute{Universit\"at Hamburg, Hamburg, Germany \and
		Federal Statistical Office (Destatis), Wiesbaden, Germany
	}
	
	\maketitle
	
	\begin{abstract} 
		This work investigates the potential of Federated Learning (FL) for official statistics and shows how well the performance of FL models can keep up with centralized learning methods.
		FL is particularly interesting for official statistics because its utilization can safeguard the privacy of data holders, thus facilitating access to a broader range of data.
		By simulating three different use cases, important insights on the applicability of the technology are gained.
		The use cases are based on a medical insurance data set, a fine dust pollution data set and a mobile radio coverage data set -- all of which are from domains close to official statistics.
		We provide a detailed analysis of the results, including a comparison of centralized and FL algorithm performances for each simulation.
		In all three use cases, we were able to train models via FL which reach a performance very close to the centralized model benchmarks.
		Our key observations and their implications for transferring the simulations into practice are summarized.
		We arrive at the conclusion that FL has the potential to emerge as a pivotal technology in future use cases of official statistics. 
	
	\end{abstract}
	%%%%%%%%%%%%%%%%%%%%%%%%%%%%%%%%%%%%%%%%%%%%%%%%%%%%%%%%%%%%%%%%%
	
	\section{Introduction}

	The aim of national statistical offices (NSOs) is to develop, produce and disseminate high-quality official statistics that can be considered a reliable portrayal of reality~\cite{yung2022quality}.
	In order to effectively capture our rapidly changing world, NSOs are currently undergoing a process of modernization, leveraging new data sources, methodologies and technologies.
	
	NSOs have effectively extracted information from new data sources, 
	such as scanner data\footnote{Scanner data in consumer price statistics and for determining regional price differences \url{https://www.destatis.de/EN/Service/EXSTAT/Datensaetze/scanner-data.html}, accessed on July 17, 2023}
	or Mobile Network Operator (MNO) data\footnote{Use of MNO data \url{https://cros-legacy.ec.europa.eu/content/12-use-mno-data_en}, accessed on July 17, 2023}.
	However, the potential of numerous other data sources, including privately held data\footnote{Guidance on private sector data sharing
	\url{https://digital-strategy.ec.europa.eu/en/policies/private-sector-data-sharing}, accessed on July 17, 2023} or data from certain official entities, remains largely untapped.
	Legal frameworks, which are fundamental to official statistics, only adapt slowly to changing data needs and currently hinder access to valuable new data sources.
	Cooperation with potential data donors faces restrictions due to concerns about privacy, confidentiality, or disclosing individual business interests.

	In the meantime, the methodology employed by NSOs is evolving, with machine learning (ML) gaining substantial  popularity and, as a result, undergoing a process of establishment.
	ML has been applied in various areas of official statistics (e.g.~\cite{dumpert2017einsatz, beck2018official,unece2022official}), 
	and new frameworks such as~\cite{yung2022quality} address the need to measure the quality of ML.
	
	Within official statistics, ML tools have proven effective in processing new data sources, such as text and images, or enabling the automation of statistical production tasks, including classifying information or predicting not (yet) available data.

	\textbf{Federated learning (FL)} is an emerging approach within ML that provides immense unexplored potential for official statistics. It addresses the challenge of extracting and exchanging valuable global information from new data sources without compromising the privacy of individual data owners. 
	Introduced in~\cite{mcmahan2017communication}, FL enables collaborative model training across distributed data sources while preserving data privacy by keeping the data localized.
	In scenarios where external partners are unwilling to share individual-level information due 
	to regulatory or strategic
	considerations, but still aim to analyze or disseminate global insights in their field of application, NSOs can offer trustworthy solutions by utilizing FL.
	In return, FL empowers contributing NSOs to 
	integrate new data sources into statistical production.
	
	Although FL has been successfully applied to many domains, to the best of our knowledge, besides our work only one currently presented study investigates the applicability of FL to the field of official statistics. 
	In a proof of concept (PoC) by the United Nations (UN), FL is applied to estimate human activity based on data collected from smart and wearable devices~\cite{petscan, unstats}. 
	The PoC emphasizes operative aspects of FL coordinating multiple NSOs and benefits of additional privacy enhancing technologies.	
	
	The main contribution of this paper lies in presenting three additional applications of FL that address current data need representative for official statistics. 
	Complementary, we emphasize measuring the numerical predictive performance and reproducibility by openly sharing our code, which, in two instances, is applied to publicly available data. 
	In the first simulation related to health, individual healthcare costs are predicted utilizing tools for regression.
	In the second simulation related to \textbf{sustainability}, current fine dust pollution levels are classified based on meteorological data.
	In the third simulation related to \textbf{mobility}, the daily range of movement of mobile phone users are classified by MNO data.
	The first two simulations focus on assessing the estimation performance achieved by FL in comparison to centralized models that have complete access to all available data. 
	The third application presents valuable insights and lessons learned from the implementation of FL, involving the active participation of a real external partner. 
	We draw conclusions on the applicability of FL in NSOs in~\autoref{sec:implications-off-statistics}, which are summarized in \autoref{sec:conclusion}.
	
	\section{Background}
	\label{sec:background}
	
	Before presenting the simulated use cases in \autoref{sec:simulations}, this section provides an overview of FL and privacy challenges with ML.
	
	\subsection{Federated Learning}
	\label{subsec:fl}
	In FL, a centralized server (or aggregator, in our case a NSO) coordinates the process of training a ML model (mainly deep neural networks) by initializing a global model and forwarding it to the data owners (clients).
	In each training round, each client trains the model with their private data and sends the resulting model back to the central server.
	The central server uses a FL method to aggregate the updates of the participants into the next iteration of the global model and starts the next round by distributing the updated model to the clients.
	This process is repeated to improve the performance of the model.
	%, just as in a regular ML training process.
	
	NSOs primarily strive to generate global models that accurately represent the available data, which, in our setting, is distributed among multiple clients. Thus, we compare the performance of FL to models with access to the combined data of all clients.
	Alternatively, if upcoming applications seek to supply each client with an optimized individual model by leveraging information from the other clients, \emph{personalized} FL can be used. This approach is not covered in this paper but can be found in ~\cite{kulkarni2020survey, hu2018federated}.
	
	\subsection{Privacy Challenges with Machine Learning}
	\label{subsec:privacy-background}
	
	When training data for a ML model is distributed among multiple parties, the data traditionally needs to be combined on a central server prior to training an ML model.
	FL has become a popular alternative to this approach, as it allows to train a model in a distributed way from the start, without the need to aggregate training data first.
	Thus, using FL has the privacy advantage that there is no need to exchange private training data.
	Instead, data holders can train a global model collaboratively in a distributed fashion, without transferring any data record.
	
	But although FL makes sharing private training data obsolete,
	there are other privacy challenges inherent to ML which have also been observed for FL.
	While ML models are always trained to fulfill a dedicated task, often more information than strictly necessary for fulfilling the task is extracted into the model weights during training~\cite{song2017machine}.
	This excessive, and potentially private, information in the model weights is called privacy leakage.
	In general, this leakage can be leveraged by any party who has full access to a model and its trained weights.
	
	One concrete example of such a privacy attack is \emph{training data extraction}~\cite{zhu2019deep}, which allows extracting data records from a trained model.
	Another known attack is	\emph{model inversion}~\cite{hitaj2017deep}, where repeated requests to the model are used to reconstruct class representatives. 
	\emph{Membership inference}~\cite{shokri2017membership}
	aims at individual training data records: the attack's target is to decide whether a specific data record was part of the training data.
	Building on the original proposal, other works have transferred membership inference attacks to the FL scenario~\cite{nasr2019comprehensive}.
	Last but not least, \emph{property inference} attacks~\cite{melis2019exploiting} allow to deduce statistical properties of the target model's training data.
	This is especially relevant in FL scenarios, where the characteristics of each client's local data set can be highly sensitive, e.g., in medical domains.
		
	The applicability of these attacks depends on the concrete use case, the type of model and other factors.
	Concerning attacker models, i.e., the scenario in which an attack is executed, some FL-specific attacks rely on a malicious aggregator.
	Nonetheless, all attacks mentioned above also work in an environment where not the aggregator, but one of the FL clients is the attacker.
	Hence, even if the aggregator can be trusted, e.g., because the aggregator's role is assumed by a NSO, these attacks can still be executed by other FL clients.
	Analyzing the individual privacy leakage of the simulated use cases in this paper are out of scope.
	Nonetheless, raising awareness to these issues, e.g., by communicating potential risks to clients in an FL scenario, should not be neglected.
	Beyond this, strategies under the umbrella term \emph{privay-preserving machine learning}~(PPML) can help to mitigate these risks~\cite{yin2021comprehensive}.

	\subsection{Frameworks}
	In our simulations, we use the frameworks TensorFlow\footnote{TensorFlow \url{https://www.tensorflow.org/}, accessed on July 17, 2023} for neural networks 
	and TensorFlow Federated\footnote{TensorFlow Federated: Machine learning on decentralized data \url{https://www.tensorflow.org/federated}, accessed on July 17, 2023} for FL.
	We use PyCaret\footnote{PyCaret \url{https://pycaret.org/}, accessed on July 17, 2023} for automizing benchmark experiments in the centralized settings and scikit-learn\footnote{scikit-learn, Machine Learning in Python \url{https://scikit-learn.org/}, accessed on July 17, 2023} for data processing.

	The code we have written for this work is openly available on GitHub\footnote{Code repository for this paper: \url{https://www.github.com/joshua-stock/fl-official-statistics}, accessed on July 17, 2023. Note that for the mobile radio coverage simulation, the code has only been executed locally on the private data set, hence it is not included in the repository.}.
	
	\section{Simulations}
	\label{sec:simulations}
	
	Most relevant for NSOs is \emph{cross-silo} FL, where a few reliable clients train a model, e.g.\ official authorities.
	In contrast, \emph{cross-device} FL uses numerous clients, e.g.\ smartphones, to train a model.
	To analyze the potential of cross-silo FL for official statistics, we run simulations with three different data sets. 
	For each use case, we first compute benchmarks by evaluating centralized ML models, i.e., models which are trained on the whole data set.
	Afterwards, we split the data set and assign the parts to (simulated) FL clients for the FL simulation.
	This way, we have a basis for interpreting the performance of the model resulting from the FL training simulation.
	The performance metrics of the trained ML models (including coefficient of determination \rsquared{} or accuracy) are computed on test sets of each data set.

	\subsection{Medical insurance data}
	\renewcommand{\arraystretch}{1.2} % increase vertical distance between lines
	\setlength{\tabcolsep}{0.15cm}
		The demand for timely and reliable information on public health is steadily increasing.
		The COVID-19 pandemic has significantly accelerated this trend, raising questions about the financial feasibility of our healthcare system and the availability of medical supplies.

		Thus, our first experiment focuses on modeling a regression problem related to healthcare by considering the following question: 
		Given an individual's health status characteristics, what is the magnitude of their insurance \emph{charges}?
        We aim to address two primary questions. 
        Firstly, we explore the suitability of neural networks in comparison to other models for the regression task. 
        Secondly, we assess the feasibility of utilizing a simulated decentralized data set in an FL setting to tackle the problem.
	
	\paragraph{Data set}
	The given data set links medical insurance premium \emph{charges} 
to related individual attributes\footnote{US health insurance dataset \url{https://www.kaggle.com/datasets/teertha/ushealthinsurancedataset}, accessed on July 17, 2023}. 
	Considered are the six features \emph{age}, \emph{sex}, \emph{bmi} (body mass index), \emph{children} (count), \emph{smoker} (yes/no) and four \emph{regions}.
	In our studies, the feature \emph{region} was excluded during FL training and solely utilized for partitioning the data within the FL setting.
	In total, the data set consists of \num{1338} complete records, i.e.\ there are no missing or undefined values.
	Also, the data set is highly balanced: The values in \emph{age} are evenly dispersed, just as the distribution of male and female records is about 50/50 (attribute \emph{gender}) and each \emph{region} is represented nearly equally often.
	The origin of the data is unknown, however its homogeneity and integrity suggest that it has been created artificially.

	\paragraph{Data preprocessing}

	We encode the binary attributes \emph{sex} and \emph{smoker} into a numeric form (0 or 1).
 	The attributes \emph{age}, \emph{bmi} and \emph{children} are scaled to a range from 0 to 1.
	In the centralized benchmarks, the attribute \emph{region} is one-hot-encoded.

	\paragraph{Setup}

        We aim to investigate the suitability of neural networks for estimating insurance \emph{charges} and explore the extent to which this problem can be addressed using a FL approach. To achieve this, we compare different models and evaluate their performance.

        A basic fully connected neural network architecture, that takes five input features, is utilized. The network consists of three hidden layers with 40, 40, and 20 units in each respective layer. 
        Following each layer, a Rectified Linear Unit (ReLU) activation function is applied. 
        The final output layer comprises a single neuron. 
        To optimize the network, the Adam optimizer with a learning rate of 0.05 is employed.
        In the federated setting, we utilize the same initial model but integrate FedAdam for server updates. 
        This decision is based on previous research~\cite{reddi2020adaptive}, which emphasizes the benefits of adaptive server optimization techniques for achieving improved convergence.
        
        In the centralized approach, we allocate a training budget of 100 epochs. 
        In contrast, the federated approach incorporates 50 rounds of communication between the client and server during training. 
        Each round involves clients individually training the model for 50 epochs.
        To track the running training, $10\%$ evaluation data is used by each client in the FL setting and $20\%$ is used in the centralized scenario. 
        It is neglected in calculating the final test performance.
        The remaining shallow learning models undergo hyperparameter optimization using a random search approach with a budget of 100 iterations. 
        We evaluate all models using 5-fold cross validation.

    \begin{table}[htbp]
      \centering
      \renewcommand{\arraystretch}{1.5} % Adjust the spacing between rows
      \begin{tabular}{lcc} % Added vertical line after the first column
        \toprule
        \multirow{1}{3.1cm}{\textbf{Model}} & 
		\multirow{1}{2.2cm}{\centering\textbf{\rsquared ($\pm$ std)}} & 
		\multirow{1}{*}{\centering\textbf{Rel. loss (\%)}} \\
        \midrule
        neural network    & $81.5 (4.01)$  & 3.5 \\
        neural network (federated)   & $78.4 (3.13)$  & 7.2 \\
        random forest    & $84.5 (4.73)$ & 0.0\\
        XGBoost   & 84.3 (3.96) & $0.2$    \\
        decision tree    & $84.1$ (4.23) & $0.5$    \\
        k-nearest neighbors   & $74.4$ (5.53) & $12.0$    \\
        linear regression    & $72.8$ (6.07)& $13.8$    \\
        \bottomrule
      \end{tabular}
      \linebreak
      \caption[Model performance comparison.]{Performance comparison of different prediction models for the medical insurance use case.
      The performance is quantified using \rsquared\ in \%, along with the corresponding standard deviation (std).
      Additionally, the relative loss to the best centralized model (rel.\ loss) is reported.}
      \label{tab:med-perf}
    \end{table}

        \paragraph{Results}
        We conduct a performance comparison of the models based on their 5-fold cross-validation \rsquared{} scores and consider their standard deviation (see \autoref{tab:med-perf}). 
        The random forest model achieves the highest performance with an \rsquared{} of 84.5~\%, closely followed by XGBoost and Decision Tree, which scores 0.2 and 0.5 percentage points lower, respectively.

        The neural network model achieves an \rsquared{} of 81.5~\%, indicating a performance 3.5~\% worse than the best model. 
        However, it still provides a reasonable result compared to K-Nearest Neighbors~(KNN) and Linear Regression, which obtain significantly lower \rsquared{} scores of 12~\% and 13.8~\%, respectively.
        
        The Federated neural network demonstrates an \rsquared{} of 78.4~\%, slightly lower than the centralized neural network but 7.2~\% worse than the random forest model. 
        Notably, the Federated neural network exhibits a lower standard deviation of 3.99 compared to the centralized neural network (4.92) and also outperforms the random forest model (4.73) in this regard.

	\paragraph{Discussion}

        Based on the research questions, we can draw clear conclusions from the findings presented in \autoref{tab:med-perf}.
        Initially, we compared the performance of different models, including a simple neural network. Although the random forest model outperformed others, its performance was only 3.5~\% higher, distinguishing it significantly from models such as KNN and linear regression, which performed 12~\% and 13.8~\% worse than the random forest, respectively.
        
        The observed performance decrease from 81.5~\% to 78.4~\% in the FL approach can be attributed to the training process and falls within a reasonable range. 
        Considering the privacy advantages of FL, the 7.2~\% accuracy loss compared to the best model is acceptable, particularly when taking into account the reduction in standard deviation from 4.92 to 3.99.
        
        Although this example is hypothetical, it highlights the potential benefits and importance of FL in official statistics. It showcases how FL provides access to crucial data sets for ML while maintaining nearly negligible loss in accuracy compared to a centralized data set.

	\subsection{Fine dust pollution}
	
	Reducing air pollution is a significant part of the Sustainable Development Goals (SDGs) established by the United Nations\footnote{Air quality and health \url{https://www.who.int/teams/environment-climate-change-and-health/air-quality-and-health/policy-progress/sustainable-development-goals-air-pollution}, accessed on July 17, 2023}. To measure progress toward achieving SDGs, NGOs and other data producing organizations developed a set of 231 internationally comparable indicators, including \emph{annual mean levels of fine particulate matter (e.g. PM$_{2.5}$ and PM$_{10}$)}.
	\cite{hu2018federated} showed that personalized FL can be used to extract timely high frequent information on air pollution more accurately than models using centralized data.
	
	\begin{figure}
		\centering
		\includegraphics[width=0.5\textwidth]{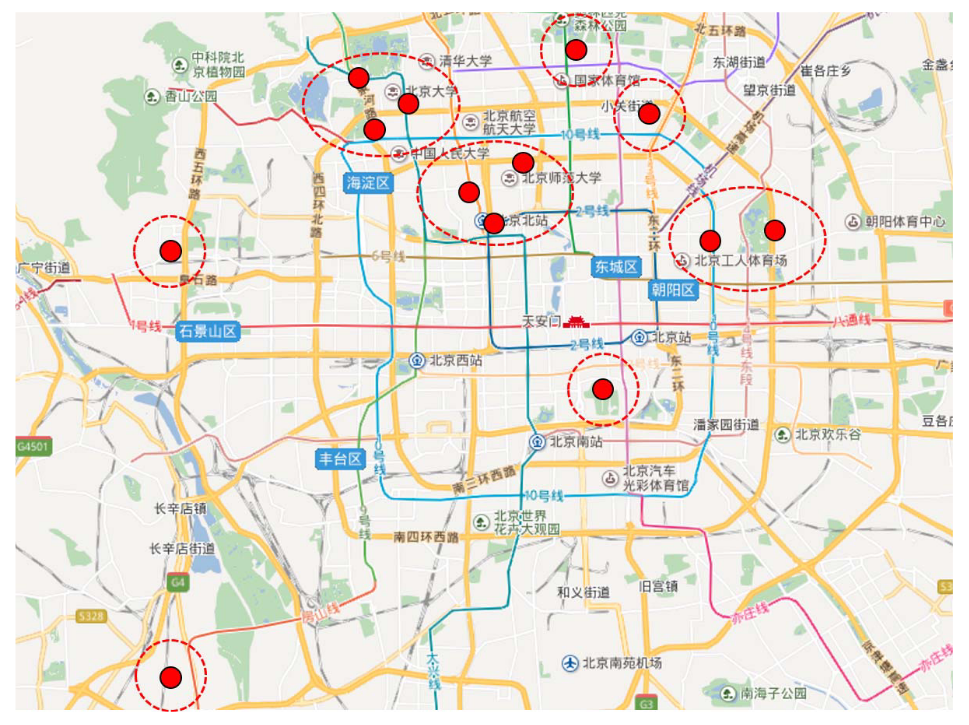}
		\caption{Location of meteorological stations for the fine dust pollution simulation on a map of Beijing, China. 
			12 of the 13 stations are included in the public data set which we have used for our simulations.
			The dashed lines mark \emph{regions} of the \enquote{Region-Learning} approach in~\cite{hu2018federated}. 
			Image source:~\cite{hu2018federated}.}
		\label{fig:pl-beijing-stations}
	\end{figure}

	In our second use case, we provide a comparison between centralized and FL models (without personalization) and make the developed code and methods accessible. 
	It should be noted that we utilize a slightly different data set and methodology compared to~\cite{hu2018federated}, which we explain at the end of this section.
	We model a classification task in which the current fine dust pollution is inferred based on meteorological input data.
	More precisely, 48 consecutive hourly measurements are used to make a prediction for the current PM$_{2.5}$ pollution (the total weight of particles smaller than \SI{2.5}{\micro\meter} in one \SI{}{\cubic\metre}).
	The output of the predictor is one of the three classes \emph{low}, \emph{medium} or \emph{high}.
	The thresholds for each class are chosen in a way such that the samples of the whole data set are distributed evenly among the three classes.
	
\begin{figure}
	    \centering
	    \includegraphics[width=0.9\columnwidth]{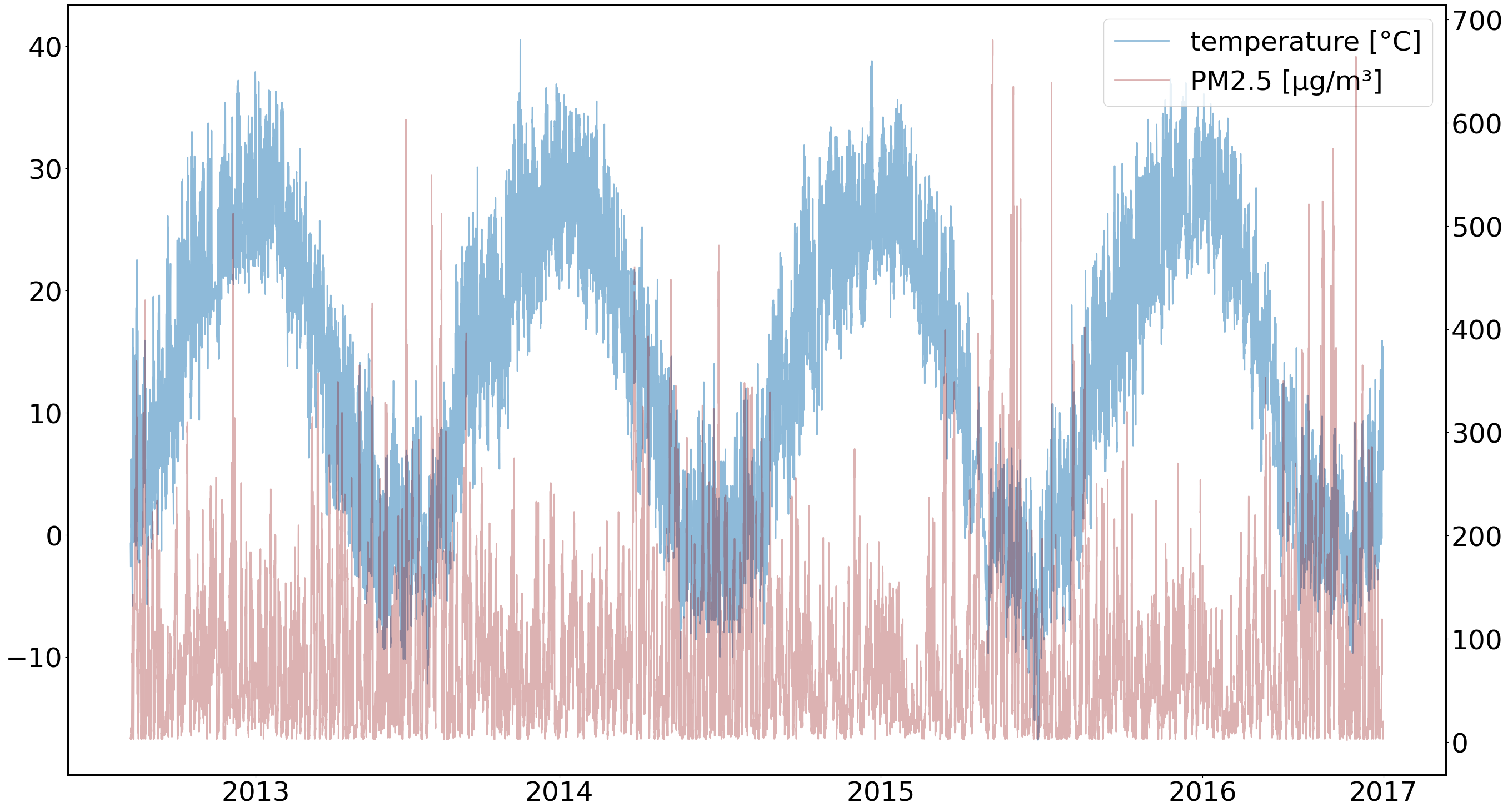}
	    \caption{Example plot for the data of one meteorological station and the two features PM$_{2.5}$ and temperature. The four-year time span is clearly visible by the temperature wave, due to hot summers and cold winters.}
	    \label{fig:pl_data}
	\end{figure}

	\paragraph{Data set}
	The data set we use is a multi-feature air quality and weather data set~\cite{zhang2017cautionary} which is publicly available online\footnote{Beijing multi-site air-quality data set \url{https://www.kaggle.com/datasets/sid321axn/beijing-multisite-airquality-data-set}, accessed on July 17, 2023}.
	It consists of hourly measurements of 12 meteorological stations in Beijing, recorded over a time span of 4 years (2013--2017).
	\autoref{fig:pl-beijing-stations} depicts the locations of the 12 stations in Beijing.
	In total, more than \num{420000} data records are included in the data set.
	Although some attributes are missing for some data records, most records have data for all the 17 attributes.
	An example plot for the two attributes PM$_{2.5}$ and temperature is shown in \autoref{fig:pl_data}.

	\paragraph{Data preprocessing}
	To complete the missing data records, we use linear interpolation.
	We apply one-hot encoding to the wind direction attribute.
	All other features are scaled to a range from 0 to 1.
	For the attributes PM$_{10}$, SO$_2$, NO$_2$, CO and O$_3$, 
	we observe a high correlation with the target attribute and thus exclude them from training.
	80\% of the data are used as training data, the rest is used as test data.

	\paragraph{Setup}
	As in the first use case, we implement a centralized learning benchmark and compare it with a FL approach.
	We model one FL client per meteorological station and split the data accordingly, while the benchmark model is trained with data from all 12 stations.
	In both settings, we use neural networks with LSTM (long-short term memory) layers and apply 5-fold cross validation.
	The architecture of the neural networks is similar across both settings and has been manually tuned to reach a good performance:
	The input layer is followed by a 10-neuron LSTM layer, a dropout layer with a dropout rate of 25\%,
	a 5-neuron LSTM layer, another dropout layer with a dropout rate of 35\% and a 3-neuron dense
	layer for the classification output.
	For the same reasons as in the first use case, we use the Adam optimizer and apply a learning rate of 0.05 on the server and 0.005 on the client.
	The client learning rate is decreased every 64 epochs by a factor of 10 to facilitate fine-tuning in later stages of the training.
	The total training budget we have allocated is 10 epochs for centralized learning and 200 epochs for FL (with a single round of local training per epoch).
	
	\paragraph{Results}
	A summary of our results for the fine dust pollution use case is provided in \autoref{tab:pm-results-table}.
	Depicted are the means of our 5-fold cross validation experiments.
	
	The centralized learning benchmark reaches a mean classification accuracy of 72.4\%, with similarly high numbers for precision and recall (72.8\%, respectively 72.3\%).
	In comparison, the FL classifier reaches a performance of both an accuracy and a recall of 68.0\% and a precision of 67.9\%.
	The relative standard deviation is higher in the FL scenario for all three metrics, reaching from +2.67 percentage points (accuracy) to +2.9 percentage points (both precision and recall).
	
	An exemplary confusion matrix for one of the five resulting models of the centralized learning is depicted in \autoref{tab:pl-global-conf}.
	Most misclassifications are made for the \emph{medium} class.
	The same could be observed for the other models (both in centralized and federated learning).
	
	\begin{table}[]
		\centering
		\begin{tabular}{lcccc} 
			\toprule
			\textbf{Model} & \textbf{Accuracy ($\pm$ std)} & \textbf{Precision ($\pm$ std)} & \textbf{Recall ($\pm$ std)}  & \textbf{Rel. loss (\%)}\\ 
			\midrule
			neural network & 72.4\% (4.92)   & 72.8\% (8.66) & 72.3\% (8.10) & 0.0    \\ 
			neural network (fed.)   & 68.0\% (7.59) & 67.9\% (10.05) & 68.0\% (9.59) & 5.9-6.7  \\
			\bottomrule
		\addlinespace[4pt]
		\end{tabular}
		\caption{Performance in the fine dust pollution simulation. The span of the relative loss refers to all three metrics.}
		\label{tab:pm-results-table}
	\end{table}
	
	\begin{table}
		\small
		\centering
		\begin{tabular}{lccc}
			\toprule
			true class / \textbf{predicted class} & \textbf{low} & \textbf{medium} & \textbf{high} \\
			\midrule
			low & \num{114450}  & \num{23712} & \num{5769} \\ 
			medium & \num{23635} & \num{80718} & \num{33754} \\
			high & \num{1944} & \num{24629} & \num{111581} \\
			\bottomrule
			\addlinespace[4pt]
		\end{tabular}
		\caption{Exemplary confusion matrix for one of the five models in the cross-validation training of the centralized model for the fine dust pollution use case.}
		\label{tab:pl-global-conf}
	\end{table}
	
	\paragraph{Discussion}

	Compared to the first use case, the training database is significantly larger.
	With 12 clients, there are also four times as many participants in the FL scenario as in the first use case.
	Still, the performance decrease is small, with an accuracy of 68.0\% (FL) compared to 72.4\% in the centralized training scenario.
	
	Apart from preprocessing the data set, 
	another time-consuming part of the engineering was tuning the hyperparameters of the FL training.
	Tools for automatic FL hyperparameter optimization were out of scope for this work, thus it was necessary to manually trigger different trial runs with varying hyperparameters.

	\paragraph{Comparison with literature}
	The authors of~\cite{hu2018federated} compare the results of their personalized FL strategy \enquote{Region-Learning} to a centralized learning baseline and standard FL.
	Although according to the authors, their personalized FL approach outperforms the other two approaches (averaged over the regions by 5 percentage points compared to standard FL), 
	we want to stress that Region-Learning has another goal than standard FL -- namely multiple specialized models, and not one global model as in standard FL and most use cases for official statistics (also see \autoref{subsec:fl}).
	
	Furthermore, \citeauthor{hu2018federated} have not provided sufficient information to retrace their experiments.
	Especially the number of classes for PM$_{2.5}$ classification and information on the features used for training the classifiers are missing, so that their results are hard to compare to ours.
	For example, setting the number of classes to 2 and using all features of the data set (including the other pollution attributes PM$_{10}$, SO$_2$ etc.) would significantly ease the estimation task.
	Also, we have no information on whether cross validation was applied in the work of~\citeauthor{hu2018federated}.
	Two more hints in the paper~\cite{hu2018federated} suggest that they have used a slightly different data set than we have: The data set they describe includes \enquote{more than \num{100000}} data records from 13 meteorological stations in Beijing, while our data set contains more than \num{420000} records from 12 stations.

	One consistency across both their work and ours is the accuracy drop from centralized learning to FL, with 4 percentage points in~\cite{hu2018federated} and 4.4 percentage points in our work.
	
	\subsection{Mobile radio (LTE)}
	Mobile Network Operator (MNO) data is a valuable source for obtaining high-frequency and spacial insights in various fields, including population structure, mobility and the socio-economic impact of policy interventions. However, a lack of legal frameworks permitting access to data of all providers, as seen in cases like Germany, constrain the quality of analysis~\cite{mobilfunkdaten}.
	Accessing only data of selected providers introduces biases, making FL an attractive solution to enhance the representativeness by enabling the aggregation of insights from multiple major MNOs.

	Thus, our third use case is based on private MNO data owned by the company umlaut SE\footnote{umlaut website \url{https://www.umlaut.com/}, accessed on July 17, 2023}. Different from the first two use cases, we had no direct access to the data, just as the aggregation party in realistic FL settings.
	While this allows for practical insights, it also comes with constricted resources in the private sector.
	Hence, the focus of this use case is more on practical engineering issues of FL and less on optimal results.
	
	The data set contains mobile communication network coverage data, including latency and speed tests, each linked to the mobile LTE devices of individual users and a specific timestamp.
	The data records are also associated with GPS coordinates, such that a daily \enquote{radius of action} can be computed for each user.
	This radius describes how far a user has moved from their home base within one day.
	The user home bases have also been computed on the available data -- a home base is defined as the place where most data records have been recorded.
	The ML task we model in this use case is to estimate the daily radius of action for a user, given different LTE metrics of one particular day (see below).
	
	\paragraph{Data set}
	The whole data set originally contains \num{286329137} data records.
	The following features of the data set have been aggregated for each day and user: \emph{radius of action} in meters, \emph{share of data records with Wi-Fi connection} and the variance and mean values for each of the following LTE metrics: \emph{RSRQ}, \emph{RSRP}, \emph{RSSNR} and \emph{RSSI}. 
	The date has been encoded into three numeric features (\emph{calendar week}, \emph{day of the week} and \emph{month}) and the boolean feature \emph{weekend}.
	
	\paragraph{Data preprocessing}
	We set a specific time frame of six months and a geofence around the German state of North Rhine-Westphalia.
	All other records are excluded -- leaving \num{2718416} records in the data set.
	Additionally, we apply a filtering strategy to clean our data: each user in the database needs to have data for at least 20 different days (within the time span of six months) and 10 records on each of these days.
	Otherwise, all records of this user are discarded.
	After the second filtering step, there are \num{1508102} data records in the data set.
	We scale each feature to a range from 0 to 1 and then use for training, validating and testing our models.
	
	60\% of the data are used as training data, 20\% are used as validation data and the remaining 20\% as test data.
	For FL, we have divided the data set according to the mobile network operators (MNOs) of the users.
	Since more than 99.6\% of the data records are associated with three major providers, the other 0.4\% of the data records (belonging to 29 other MNOs) are eliminated from the data set.	
	
	\paragraph{Setup}
	We use two centralized learning benchmarks: a random forest regressor and a neural network, which have both been subject to a hyperparameter search prior to their training.
	The network architecture for both the centralized benchmark neural network and the FL training process is the same:
	The first layer consists of 28 dense-neurons and the second layer consists of 14 dense-neurons, which lead to the single-neuron output layer.
	All dense layers except for the output layer use the ReLU activation function.
	For FL, we use the SGD optimizer with a server learning rate of 3.0, a client learning rate of 0.8 and a batch size of 2.
	
	\paragraph{Results}
	The benchmarks of the centralized learning regressors are \rsquared{} values of 0.158 (random forest), 0.13 (neural network) and 0.13 (linear regression).
	For the neural network trained in the FL scenario, we achieve a slightly lower \rsquared{} value of 0.114 (see \autoref{tab:lte-results-table}).
	\begin{table}[]
		\small
		\centering
		\begin{tabular}{lcc}
			\toprule
			\textbf{Model} & \textbf{\rsquared} & \textbf{Rel. loss (\%)} \\ 
			\midrule
			neural network & 0.130 & 17.7 \\ 
			neural network (federated)  & 0.114 & 27.8 \\
			random forest & 0.158 & 0.0 \\ 
			\bottomrule
			\addlinespace[4pt]
		\end{tabular}
		\caption{Performance in the mobile radio simulation.}
		\label{tab:lte-results-table}
	\end{table}

	\paragraph{Discussion} 
	The reasons behind the weak performance of the benchmark models (\rsquared{} of 0.158 and 0.13) are not clear.
	The hyperparameters might not be optimal, since we were not able to spend many resources on hyperparameter tuning due to time constraints of the data owner.
	Another reason might be the that the modeled task (estimating the radius of action based on LTE connection data) is inherently hard to learn.
	With an \rsquared{} of 0.114, we were able to reproduce this performance in the FL setting.
	
	Since the private data set in this use case has not left company premises, there are important lessons to be learned from a practical perspective:
	\begin{enumerate}
		\item Even if the data set is not directly available during the model engineering process, it is crucial to get basic insights on the features and statistical properties before starting the training.
		Essential decisions, such as the type of model to be trained, can be made based on this.
		\item A thorough hyperparameter optimization is needed to obtain useful results. 
		It might take a lot of time and computational resources to find hyperparameters which are suited for the task.
		\item Technical difficulties while creating the necessary APIs and setting up the chosen ML framework at the FL clients can slow down the process even more.
		Without access to the database, it might be hard to reproduce technical errors.
	\end{enumerate}
	
	While all points mentioned above were encountered in the third simulation, there was only \emph{one} party who held all data.
	In real FL scenarios with multiple data holders, the process might get much more complicated.
	
	\section{Key Observations}
	\label{sec:observations}
	Our simulations lead to the following key observations:
	
	Models trained via FL can reach a performance very close to models trained with centralized ML approaches, as we have shown in all three use cases.
	While the performance gap itself is not surprising (since the FL model has been exposed to the complete data set only indirectly), we want to stress that without FL, many ML scenarios might not be possible due to privacy concerns, trade secrets, or similar reasons.
	This is especially true for health care data, i.e., the domain of our first simulation.

	While the random forest regressor has demonstrated superior performance compared to other centralized learning benchmarks in all three simulations, exploring the potential of tree-based models within a FL context \cites{alquraan2022fedtrees, eflboost, FedGBDT} could be a promising avenue for further investigation.
	The improved interpretability and explainability over many other models, e.g., neural networks, is another advantage of tree-based models.
	
	On the other hand, random forest regressors are not suitable if tasks get more complicated.
	Also, their architecture, i.e., many decision trees which may be individually overfitted to parts of the training data, can facilitate the extraction of sensitive information of the training data and thus pose an additional privacy risk.
	
	Choosing the right hyperparameters is crucial for any ML model. 
	Since automatic HPO is still an open problem for FL algorithms, (manually) finding the right settings can be a time-consuming process.
	Developing a suitable framework for automated HPO for FL would be important future work -- although for official statistics, other issues might be more pressing at the moment (see \autoref{sec:implications-off-statistics}).
	
	In our third simulation (mobile radio data), we did not have access to the training and test data set, just like in a real-world scenario.
	This means both HPO and technical debugging needed to be performed remotely, without access to the data.
	Although this was already challenging,
	we believe that in scenarios with multiple data holders and possibly heterogeneous data sets, these tasks will be even harder.

	All FL simulations were performed on the machine which also had access to the complete data set.
	In a real-world application, where each client runs on a distinct machine, other settings and other frameworks might be more practical than TensorFlow Federated.

	Last but not least, we want to emphasize that FL, despite its privacy-enhancing character, 
	may still be vulnerable to some ML privacy issues (see \autoref{subsec:privacy-background}).
	Hence, analyzing and communicating these risks is an important step before an application is rolled out in practice.
	
	\section{Implications for Official Statistics}
	\label{sec:implications-off-statistics}
	% intro
	In this work, we have demonstrated how FL can enable NSOs to address pressing data needs in fields that are relevant to policymakers and society. 
	Official statistics are characterized by high accuracy while underlying strict standards in confidentiality and privacy. 
	Accuracy, explainability, reproducibility, timeliness, and cost-effectiveness are essential quality dimensions for statistical algorithms~\cite{yung2022quality}. 
	In this setting, our findings indicate that FL bears significant potential to support statistical production and improve data quality.
	
	% advantages
	We have shown that FL can empower NSOs to generate reliable models that accurately capture global relations. 
	In each of our use cases, the FL-generated models exhibited nearly identical predictive performance compared to a model created by combining all available data. 
	Each model architecture that performed well on centralized or local data could be easily adapted to a FL training process with a similar level of predictive performance only using distributed data.
	
	% personalization
	If upcoming applications require to optimize an individual model for each participating party, personalized FL can be used to generate potentially improved models tailored to individual clients. 
	This increases the interest to cooperate for each participating party, as it offers 
	to enhance the analytic potential for each client and the server.
	However, it is important to note that this customization may come at the cost of global predictive performance. 
	
	%privacy and trustworthiness
	FL provides the main advantage of not needing to exchange sensitive data (see \autoref{subsec:privacy-background}). 
	Additionally, there is no need to store or process the complete data set centralized in the NSOs.
	
	% relation to legislation
	NSOs can be empowered to appraise novel data sources sans the need for new legislation. In cases where legislative changes prove impractical, FL provides a crucial pathway to assess and prepare for regulations' modernization. 
	By showcasing the advantages and implications of accessing new data sources before legal frameworks permit, FL not only significantly accelerates and relieves statistics production but also occasionally enables it.
	
	% further work: technology
	To ensure successful future implementations of FL in NSOs, it is essential to focus on further advancements. Specifically, improvements in communication frequency are crucial to enable high-speed and efficient exchanges. Our observations indicate that FL generally requires a greater number of epochs (distributed across communication rounds) compared to centralized training to achieve similar performance levels. In our use cases, even with small datasets, we found that at least 50 rounds of communication were necessary. In real-world applications, this would result in high delay and cost. Therefore, the development of infrastructure for seamless sending and receiving ML models is necessary. Addressing this challenge, we discovered that the implementation of adaptive server optimization techniques reduced the training rounds and contributed to training stability. As a result, we recommend the use of adaptive optimizers to help minimize communication costs and enhance the efficiency of FL processes. By incorporating such adaptive optimization methods, NSOs can optimize the performance and effectiveness of FL while reducing the burden of communication overhead. 
	
	% further work: partners
	Additionally, it is crucial to provide partners with the necessary tools to update models effectively. 
	This requires coordination of the server and expertise from all participating parties. In practice, real-world applications of FL often involve the challenge of harmonizing client data without directly accessing it. Achieving an optimized model architecture uniformly across all clients also necessitates the knowledge and collaborative efforts of the clients themselves. 
	Providing comprehensive tools and resources to partners enables them to actively contribute to the model updating process while maintaining data privacy and security. 
	
	% future of FL
	FL is evolving rapidly and both industry and research will continue to improve the field in the coming years. The performance and efficiency of practical FL frameworks is expected to be further optimized. Similarly, we expect the development of more usable PPML algorithms including the ones based on Secure Multi-Party Computation (SMPC) and Homomorphic Encryption (HE) – allowing for provably secure collaborative ML. Although such PPML methods have been proposed and frameworks exist, their performance today is often far from acceptable for many practical applications. With more standardization and simpler, respectively more efficient, applications, FL will become even more beneficial to official statistics.
	
	% Summary
	In summary, FL should indeed be recognized as an important technology that can facilitate the modernization of legal frameworks for official statistics. 
	It enables NSOs to safely use publicly relevant information that is not expected to be accessed by future legal frameworks, ultimately enhancing the quality and relevance of official statistics. 
	However, further development is still required to fully realize the potential of FL in this context.

	\section{Conclusion}
	\label{sec:conclusion}

	%FL has a lot of potential for official statistics.
	In scenarios where external partners are unwilling to share individual-level information but still aim to analyze or disseminate global insights in their field of application, FL can help to overcome these issues.
	We have shown across a range of three simulated use cases that FL can reach a very similar performance to centralized learning algorithms.
	Hence, our results indicate that if classic (centralized) ML techniques work sufficiently well, FL can possibly produce models with a similar performance.
	
	One of the next steps to transfer FL into the practice of official statistics could be to conduct practical pilot studies.
	These could further showcase both the applicability and challenges of FL beyond a simulated context.
	Another focus of future work in this area could be the analysis of privacy risks in FL scenarios of official statistics and potential mitigation strategies.
	This would be an important stepping stone in ensuring the privacy protection of involved parties, on top of the privacy enhancement by using FL.
	Just as in countless other domains, we expect FL to become a relevant technology for official statistics in the near future.

	\printbibliography
	
\end{document}